**Automatic Retrieval of Specific Cows from Unlabeled Videos**

J. Lyu[1], M. Ramesh[2], M. Simonds[3], J. P. Boerman[3], and A. R. Reibman[2]

[1]*Department of Computer Sciences, Purdue University, Purdue University, West Lafayette, IN USA*
[2]*Elmore Family School of Electrical and Computer Engineering, Purdue University, West Lafayette, IN USA*
[3]*Animal Sciences Department, Purdue University, West Lafayette, IN USA*

Corresponding author: Amy R. Reibman. Email: reibman@purdue.edu

*Implications*
We describe a system that retrieves video observations of Holstein cows from a continuous stream of video. The system is composed of an AutoCattloger, which builds a Cattlog of dairy cows with a single input video clip per cow, an eidetic cow recognizer which uses no deep learning to ID cows, and a CowFinder, which IDs cows in a continuous stream of video. We demonstrate its value in finding individuals in unlabeled, unsegmented videos of cows walking unconstrained through the holding area of a milking parlor. The system enables automatic continuous logging of individual cattle for weight, feed-intake, and behavior.

*Introduction*
Many methods have been proposed to recognize individual cattle from their coat patterns (Bergamini et al., 2018) or their body shape (Lassen et al., 2023). However, most use deep learning, which requires many samples to be labeled by human annotators to provide the system enough variations for robust recognition. In addition, these systems are typically evaluated using datasets, which are human curations of carefully constructed image or data samples. In contrast, we propose a system that operates on continuous video without human curation. It relies on deep learning for cow keypoint and mask localization only, and the only manual intervention is to create a single labeled exemplar video containing a new individual. This enables an automated workflow to create a video recording of each animal each day for further per-animal analysis (like body weight estimation and body condition scoring), because the system can easily add and remove cows during herd management.

*Software System*
The goal of the overall software system is to create labeled video segments that each contain a single identified cow from an input which is an unsegmented, unlabeled video containing multiple cows. There are 3 main processing components to achieve this: an AutoCattloger, an Eidetic Cattle Recognizer (ECR), and a CowFinder module. In the current system, all processing uses a top-down camera. Underlying this process is the concept of a Cattlog, which is a catalog of all the cows that might be present in any video.

The Cattlog itself contains a 2048-bit Cow Barcode for each cow; the barcode is a compact visual representation of that cow's coat pattern when viewed from above (Ramesh et al., 2023). In this work, the barcode is generated from an input image using the following steps: (a) apply a Mask-RCNN mask detector (He et al., 2020) and an HRNet keypoint detector (Sun et al., 2019), (b) apply a keypoint checker (Ramesh et al., 2023) and a keypoint rectifier (Ramesh & Reibman 2024) to the detected keypoints (shown in Fig. 1), (c) remove the background using the detected mask, (d) align the image using the keypoints to conform to a pre-determined template shape, and (e) pixelate and binarize the aligned image. An underlying principle of this approach is that deep learning is used only for keypoint localization and not for learning identities. Using this approach along with the barcode visual representation enables identification of a cow when the cow is not perfectly aligned, even with a single example image.

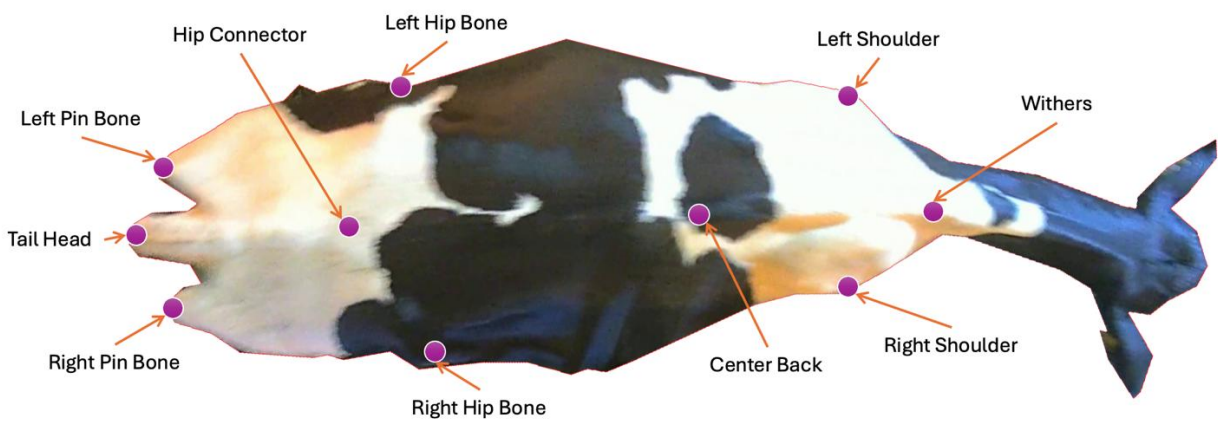

**Fig. 1.** Ten keypoints from the top view.

The AutoCattloger processes all the frames of a single video that contains a single cow to automatically build a Cattlog entry for that cow. The barcode stored in the Cattlog is the statistical mode across all processed frames. The AutoCattloger requires at least one frame in the video that clearly shows the full back of the cow. It is the only piece of the system that requires ground-truth labels.

The inputs of the ECR are an unlabeled video containing a single cow and a Cattlog. For each frame in the video, it outputs the top-1 predicted ID, the barcode of the current cow, and Hamming distances between the barcode of the current cow and the barcodes of the top-3 predicted IDs in the Cattlog. The cow ID with the smallest Hamming distance across all frames in the video is assigned as the predicted cow ID.

The CowFinder module identifies cows in a continuous stream of video. It takes as input an uncurated, unlabeled video and the Cattlog, and applies the ECR to assign a cow ID to each frame. If the Hamming distance for a given frame is larger than a manually set threshold, the predicted ID is rejected. CowFinder then clusters the results in time to obtain a start and stop time for each identified cow ID and merges any instances where the same cow was recognized more than once consecutively. This enables the creation of labeled video segments of cows from the unlabeled video.

*Video Data Collection*
Video data for this study was collected and performed at the Purdue University Animal Sciences Research and Education Center dairy unit with video data collected daily from March to April 2024 as Holstein dairy cows returned to the barn following afternoon milking. The system recorded a side-view and a top-video video using two cameras synchronized with a resolution of 1920 x 1980 at 30 fps and an encoding rate of 6000 kbps streamed using the real time streaming protocol (RTSP). Over 23 hours of video were recorded from each camera.

Each day, cows walked freely through the recording region, except those cows that were 3, 10, 17, 24, and 31 DIM. These cows were isolated to walk single file through the recording region. For each isolated cow, the ID was recorded, and the time interval during which each of these cows was visible in the side-view video was also manually recorded. These records create the ground truth for our experiment.

For the ground-truth observations, the side-view timings were used to segment the top-view videos to create so-called cut videos for each of the labeled cows. All subsequent processing used only the top-view videos. We selected one top-view cut video per cow ID as input to the AutoCattloger to form a Cattlog of 36 cows to be used for the remainder of the experiment.

*System Evaluation and Results*
We evaluated the overall system when it was applied to a continuous stream of video. We split the analysis into two pieces: the cows that walked in isolation for which we had ground truth times and IDs, and the cows that walked freely for which we did not create ground truth IDs *a priori*. Whenever the system was applied to find cows for which we did not have ground truth, a visual comparison to the Cattlog images was performed to verify the identity of each found cow.

When the cows were isolated during walking, a video segment was considered properly received if it starts and ends within the ground-truth time period. Overall, 31 of the 36 (86%) of the isolated cows (not used to form the Cattlog) were found with correct predicted cow IDs.

When cows were not isolated during walking, the system found the correct cow in 84 instances, but missed a cow 47 times when it was present in the trial, for a retrieval rate of 64%. Notably, one cow accounted for 9 of the missed instances, because the ground truth ID had been misassigned during the Cattlog process. Interestingly, the system also correctly found cows in 77 cases after they were no longer in the trial; this further demonstrates the effectiveness of the system.

*Conclusion*

We present a cow-retrieval system, which finds unlabeled cows within a continuous video stream. Unlike traditional recognition systems, it does not use deep learning for matching coat patterns. While its strength is that it uses a single observation of a cow to create the Cattlog, this also makes it more vulnerable to mistaken ID during acquisition.